\documentclass[11pt]{article}

\usepackage[preprint]{acl}

\usepackage{times}
\usepackage{latexsym}
\usepackage[T1]{fontenc}
\usepackage[utf8]{inputenc}
\usepackage{microtype}
\usepackage{inconsolata}
\usepackage{graphicx}

\usepackage[breakable]{tcolorbox}
\usepackage{amsfonts}
\usepackage{amsmath}
\usepackage{amssymb}
\usepackage{amsthm}
\usepackage{booktabs}
\usepackage{cleveref}
\usepackage{enumitem}
\usepackage{multirow}
\usepackage{soul}
\usepackage{subfig}
\usepackage{tabularx}
\usepackage{xcolor}

\usepackage{arydshln}

\newcolumntype{Y}{>{\centering\arraybackslash}X}
\newtheorem*{problem}{Problem}

\title{Repairing Regex Vulnerabilities via Localization-Guided Instructions}

\author{
  Sicheol Sung\thanks{Equal contribution.} \qquad
  Joonghyuk Hahn\footnotemark[1] \qquad
  Yo-Sub Han\thanks{Corresponding author.} \\
  Department of Computer Science, Yonsei University, Seoul, Republic of Korea \\
  \texttt{\{%
    \href{mailto:sicheol.sung@yonsei.ac.kr}{sicheol.sung},%
    \href{mailto:greghahn@yonsei.ac.kr}{greghahn},%
    \href{mailto:emmous@yonsei.ac.kr}{emmous}\}@yonsei.ac.kr}
}

\newtcolorbox{promptbox}[1][]{
  colback=gray!5,
  colframe=gray!80!black,
  coltitle=gray!30!black,
  colbacktitle=gray!30,
  boxrule=0.8pt,
  arc=2pt,
  left=6pt,
  right=6pt,
  top=4pt,
  bottom=4pt,
  fonttitle=\bfseries,
  title=Prompt,
  breakable = true,
  #1
}

\newtcolorbox{examplebox}[1][]{
  colback=gray!5,
  colframe=gray!80!black,
  coltitle=gray!30!black,
  colbacktitle=gray!30,
  boxrule=0.8pt,
  arc=2pt,
  left=6pt,
  right=6pt,
  top=4pt,
  bottom=4pt,
  fonttitle=\bfseries,
  title=Example,
  breakable = true,
  #1
}

\begin{document}
\maketitle

\begin{abstract}
Regular expressions~(regexes) are foundational to modern computing for critical
tasks like input validation and data parsing, yet their ubiquity exposes systems
to regular expression denial of service~(ReDoS), a vulnerability requiring
automated repair methods. Current approaches, however, are hampered by a
trade-off. Symbolic, rule-based system are precise but fails to repair unseen or
complex vulnerability patterns. Conversely, large language models~(LLMs) possess
the necessary generalizability but are unreliable for tasks demanding strict
syntactic and semantic correctness. We resolve this impasse by introducing a
hybrid framework, localized regex repair~(LRR), designed to harness LLM
generalization while enforcing reliability. Our core insight is to decouple
problem identification from the repair process. First, a deterministic, symbolic
module localizes the precise vulnerable subpattern, creating a constrained and
tractable problem space. Then, the LLM invoked to generate a semantically
equivalent fix for this isolated segment. This combined architecture
successfully resolves complex repair cases intractable for rule-based repair
while avoiding the semantic errors of LLM-only approaches. Our work provides a
validated methodology for solving such problems in automated repair, improving
the repair rate by 15.4\%p over the state-of-the-art. Our code is available at
\url{https://github.com/cdltlehf/LRR}.
\end{abstract}

\section{Introduction}
\label{sec:intro}

\begin{figure}[hbt]
\centering
\includegraphics[width=.98\linewidth,trim={3mm 5mm 0 0},clip]{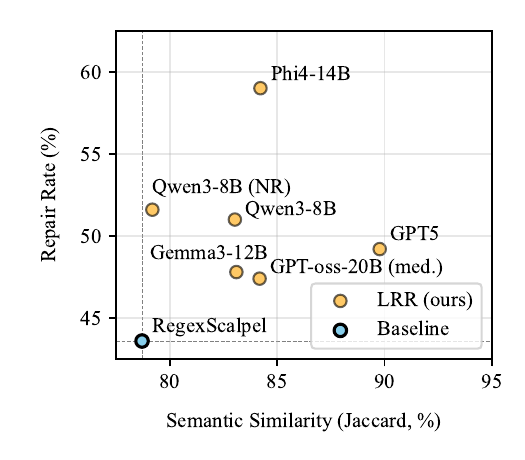}
\caption{\label{fig:performance}
    Performance comparison of the LRR framework against the baseline. The plot
    compares LRR implementations~(yellow) against the rule-based
    baseline~(blue), demonstrating successful improvement in repair
    rate~(y-axis) across semantic similarity~(x-axis).
}
\end{figure}

Regular expressions~(regexes) are a foundational tool in modern software
development, serving as the backbone for critical tasks such as input
validation, data parsing, and network security
policies~\citep{ChapmanS16,CortesM08,LiuSLGF12}. Their concise syntax and
powerful pattern-matching capabilities have made them indispensable. However,
the expressive power of modern regexes introduces a critical performance
vulnerability: certain regexes can exhibit super-linear, often exponential, time
complexity. A ReDoS attack exploits this
vulnerability~\citep{RathnayakeT14,TuronovaHHLVV22}. In such an attack, a
malicious user submits a crafted string that forces the regex engine into a
prolonged, resource-intensive matching process, leading to resource exhaustion
and service failure. The severity of this threat is evidenced by high-profile
incidents at companies like StackOverflow and Cloudflare, while numerous other
systems remain potentially vulnerable~\citep{BhuiyanCBDS24}.

The primary strategy for mitigating ReDoS vulnerabilities is the automated
repair of vulnerable regexes themselves. An alternative, replacing the
underlying regex engine, is often impractical due to high engineering costs and
the risk of introducing subtle semantic changes in matching
behavior~\citep{DavisMCSL19}. However, traditional repair approaches have
notable limitations. Synthesis-based methods require a large set of examples to
guide the repair process~\citep{LiXCCGCZ20}. In contrast, rule-based systems
rely on human-designed heuristics that offer precision and are effective at
localizing known vulnerability patterns. However, they are fundamentally
brittle; they often fail to generate semantically-equivalent and generalizable
repairs for complex vulnerability patterns.

We proposes the localized regex repair~(LRR) framework, a hybrid framework
designed to integrate the strengths of both approaches. The framework combines
the precision of symbolic analysis, which can accurately localize the vulnerable
part of a regex, with the generalization capabilities of
LLMs~\citep{WeiTBRZBYBZMCHVLDF22,ChowdheryNDBMRBCSGSSTMRBTSPRDHPBAI23,ZanCZLWGWL23}.
By pinpointing the exact source of the vulnerability, we can provide the LLM
with a precise context, allowing it to generate a targeted and semantically
correct repair. This localization-guided approach steers the LLM's powerful
generation abilities~\citep{GengDWJW24,HirschSWSBD25,OrvalhoJM25}.
\Cref{fig:performance} summarizes the increased repair rate and preserved
semantic similarity achieved by our method.

\section{Related works}
\label{sec:related}

In this section, we review existing research related to ReDoS, including methods
for vulnerability detection and mitigation, as well as approaches that utilize
symbolic analysis.

\subsection{Localization of large problems}
\label{ssec:related-localized}

Recent works have shown that tackling code-related tasks by narrowing focus to
local text, an improve both efficiency and
accuracy~\citep{ZhangZS022,PrennerR24}. For instance, \citet{TangGXSSEOKB25}
demonstrates that program repair benefits from a stages approach: the model
first attempts fixes using only the immediate local code snippet, expanding to
dependency-related context only when necessary. Similarly,
\citet{Pamies-JuarezHO13} evidences code repair can become more efficient by
relying on small, localized subsets rather than global redundancy.

In the broader LLM field, context pruning techniques~\citep{LiDGL23,HuangZC0024}
validate this strategy by showing that selectively removing redundant portions
of long documents can reduce computational cost while maintaining performance.
This established efficacy of localization in code repair strongly suggests that
precisely isolating the vulnerable sub-pattern enables robust and efficient
LLM-based regex vulnerability repair.

\subsection{In-context learning}
\label{ssec:in-context-learning}

In-context learning~(ICL) is a core capability of LLMs to learn new tasks from a
few examples provided directly in the prompt, without any weight
updates~\citep{BrownMRSKDNSSAA20,MinLHALHZ22,ShivagundeLMR24}. While early ICL
was limited to simple input-output pairs, chain-of-thought~(CoT) prompting was
introduced as a sophisticated advancement to handle more complex
reasoning~\citep{MinLZH22,Wei0SBIXCLZ22}. CoT provides an intermediate,
step-by-step reasoning process to arrive at a solution and guides the LLM to
emulate a similar deductive procedure. CoT has significantly improved
performance on intricate analytical tasks~\citep{ZhouSHWS0SCBLC23,SunLGLSGD24}.
The success of CoT has inspired more advanced variants of
CoT~\citep{WangWSLCNCZ23}, such as graph-of-thoughts~\citep{YaoLZ24} or
tree-of-thoughts~\citep{RanaldiPRRZ24} to explore reasoning paths more
effectively. Following the essence of such works, we design a CoT instruction
specifically designed to repair ReDoS.

\subsection{Vulnerability detection}
\label{ssec:related-vul-detection}

Vulnerability detection efforts include identifying common ReDoS anti-patterns
to guide developers during regex writing \cite{HassanALDS23}. Dynamic analysis
tools like ReDoSHunter \cite{LiCCXPCCC21} classify vulnerabilities using
signature-based methods for attack string generation. Further research employs
fuzzing and advanced dynamic methods to accelerate problematic string generation
and improve detection accuracy \cite{McLaughlinPSKV22,WangZLXHLYXZL23,SuHLCG24}.

Studies by \citet{SiddiqZRS24} and \citet{SiddiqZS24} show that LLMs struggle to
inherently generate ReDos-invulnerable regexes, often producing vulnerabilities
similar to those found in human code. Our approach addresses this limitation by
using a precise, rule-based module to localize the vulnerability, transforming
the problem into a manageable repair task where the LLM's generative strengths
can be effectively applied.

\subsection{Vulnerability repair}
\label{ssec:related-vul-mitigation}

FlashRegex~\citep{LiXCCGCZ20} uses programming-by-example approach to compose
ReDoS-invulnerable regexes from examples. They guarantee the invulnerability by
enforcing 1-unambiguity during synthesis. Remedy~\citep{ChidaT22} supports
symbolic regexes which allow interval character-classes instead of literals, and
extended features such as look-aheads and back-references.
RegexScalpel~\citep{LiSXCLLCCLX22} identifies and repairing only the vulnerable
spans of a given regex, minimizing semantic deviations while applying
example-based synthesis. While these synthesis-based methods are powerful, they
are often rigid and require a large set of examples to guide the repair process.
Our approach, in contrast, leverages the flexibility of LLMs to generate nuanced
and correct repairs without needing an extensive, manually-curated example.

\section{Problem definition}

\begin{figure*}
\centering
\includegraphics[width=\linewidth,trim={14mm 0 14mm 0},clip]{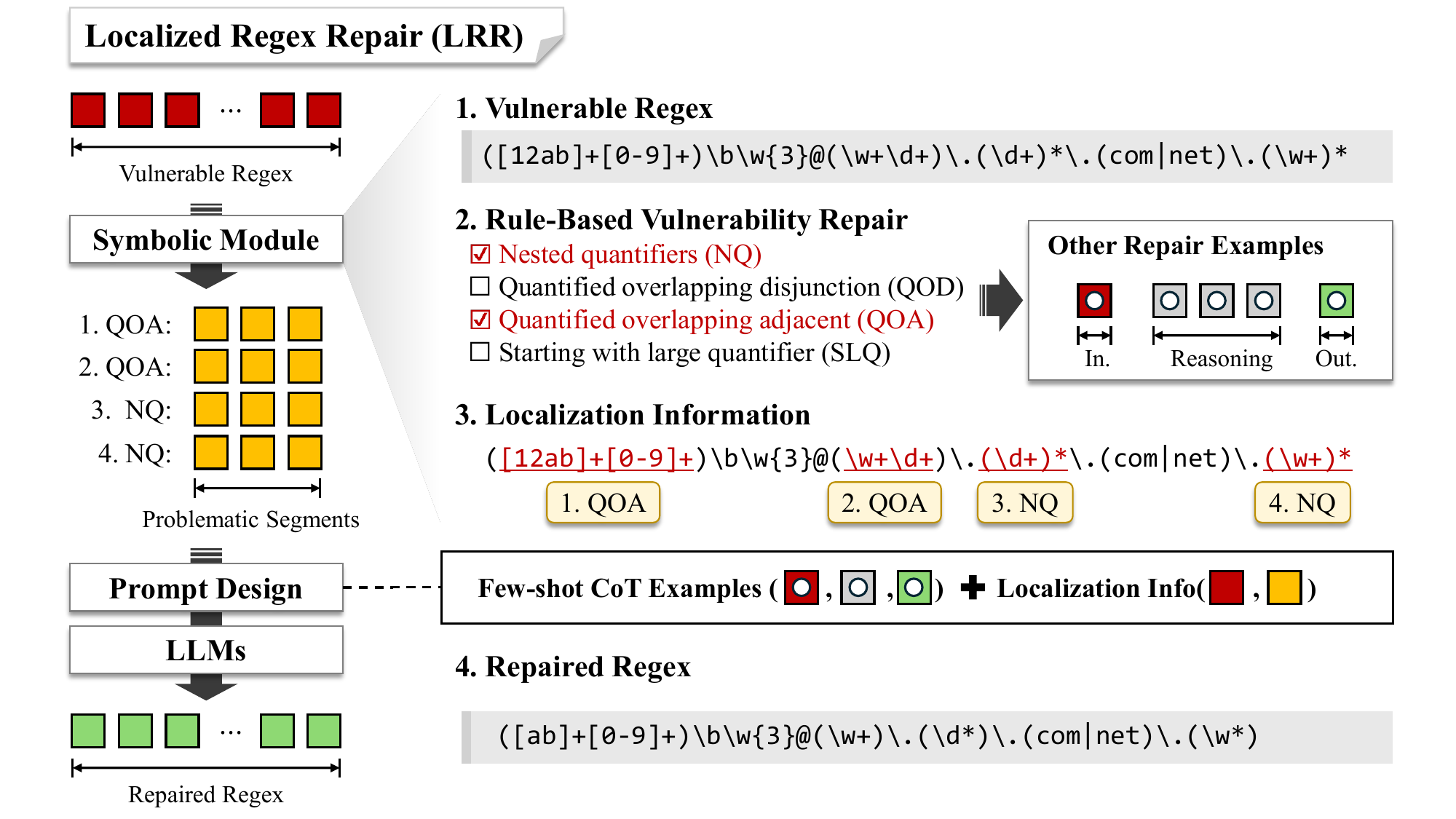}
\caption{\label{fig:framework}
    An overview of our LLR framework. Given a vulnerable regex, the symbolic
    module first analyzes and localizes the vulnerability-inducing sub-pattern.
    This identified sub-pattern is then integrated into the LLMs' prompt,
    allowing the models to focus on the problematic segment and maximize the
    repair rate.
}
\end{figure*}

The core problem we address is the automated repair of regex that are vulnerable
to ReDoS attacks. These vulnerabilities stem from catastrophic backtracking,
where certain regexes exhibit super-linear time complexity on specific inputs.
While the prevalence and danger of these vulnerabilities are well-established,
effective and scalable repair remains a significant challenge.

Traditional repair methods face critical limitations: synthesis-based approaches
require large, often unavailable, example sets, while rule-based systems are
inflexible and fail to generalize despite being precise for known patterns.

Given a vulnerable regex~$s$, our objective is to synthesize a new repaired
regex~$t$ that is safe from ReDoS attacks and semantically equivalent to $s$.
Our work operates on the hypothesis that a naive approach using an LLM is
ineffective for this task due to its difficulty in understanding the complex and
subtle causes of a vulnerability. We overcome this limitation by proposing a
hybrid framework that leverages a symbolic analysis module to precisely localize
the vulnerable sub-pattern. This process transforms a broad, complex problem
into a smaller, more manageable one, allowing the LLM to focus on generating a
targeted and correct repair.

\begin{problem}
Given a ReDoS vulnerable regex~$s$, our goal is to compose a repaired regex~$t$
such that:
\begin{enumerate}[noitemsep]
    \item ReDoS-invulnerable, and
    \item semantically and syntactically similar to $s$.
\end{enumerate}
\end{problem}

\section{Method}
\label{sec:method}

Our approach addresses the core limitation of naive LLM-based repair: their poor
performance on complex regular expressions due to a lack of precise
vulnerability analysis. We adopt a symbolic-guided in-context learning
methodology, inspired by the localize-and-fix strategy, to enable LLMs to
perform targeted and effective repairs. The LRR framework leverages the
complementary strengths of symbolic analysis and LLMs. The symbolic module
serves as a precise localization engine, identifying the exact sub-pattern
responsible for the ReDoS vulnerability. We integrate this output into a
structured prompt for the LLM. This focused context guides the LLM to
concentrate its generative power on the specific area of concern. The LLM's task
is to leverage this focused context to generate a semantically equivalent, but
ReDoS-invulnerable, repaired version of the original vulnerable regex.

The symbolic module in our framework functions as a precise vulnerability
localizer, directly leveraging the established heuristics from existing
rule-based ReDoS repair methods. This approach is based on previous works
showing that while rule-based methods are highly effective for merely detecting
ReDoS vulnerability~\citep{HassanALDS23,LiCCXPCCC21}, they often fail to repair
the detected vulnerability. Specifically, the module uses an analyzer to
pinpoint the vulnerable sub-pattern~$s'$ within the original regex~$s$
corresponding to a known ReDoS pattern.

We then teach the LLM how to utilize this localization information by providing
few-shot chain-of-thought~(CoT) examples. Each CoT example consists of the
vulnerable regex, the reasoning process and the repaired regex. The reasoning
details how the given sub-pattern~$s'$ causes the ReDoS vulnerability and how
rule-based repair methods modify it to resolve the issue. Through this approach,
we aim for the LLM to not only learn the given heuristics but also to leverage
this prior knowledge to identify vulnerable sub-patterns missed by existing
heuristics or to apply repair independently of the heuristic for the given
sub-pattern.

\section{Experimental setup}
\label{sec:experiment-setup}

We define the specific settings and evaluation criteria for the evaluation. We
use: Qwen3~\citep{QwenTeam25}, Llama3.1~\citep{Weerawardhena25},
Gemma3~\citep{Gemma25}, Phi4~\citep{AbdinABBEGHHJK24},
DeepSeek-R1-Distill-Qwen~\citep{DeepSeekAI25}, GPT-oss~\cite{OpenAI25}, and
GPT5~\cite{OpenAI25b}. We outline the specific software and hardware
specification in Appendix~\ref{app:experimental-details}.

\subsection{Prompt design}
\label{ssec:prompt-design}

We design our prompts by analyzing the successful repair processes of
RegexScalpel, a rule-based regex repair tool. The heuristics used by Scalpel are
categorized into four major vulnerability types: nested quantifiers, quantified
overlapping disjunction~(QOD), quantified overlapping adjacent and starting with
large quantifier. We convert these heuristics into a natural language
chain-of-thought reasoning process to effectively guide LLMs.

For the 5-shot prompt, we select a single example that incorporates three of the
four major repair rule categories, excluding the QOD type. Since no single
example showcases all four rule types at once, we intentionally choose a sample
that demonstrates a multi-step repair. We then add four more examples, each
showing a distinct rule category applied in isolation. This exposes the model to
a wider variety of vulnerability causes and teaches it that a regex repair might
involve multiple, successive steps. This method aids the model in learning how
to apply different rules and recognize that the repair process as a combination
of several logical steps depending on the vulnerability's complexity. The
prompts are detailed in Appendix~\ref{app:prompts}.

\subsection{Metric}
\label{ssec:metric}

The main purpose of the repair is to eliminate ReDoS vulnerabilities while
minimally transforming the original regex. Changes introduced during repair
process are either syntactic or semantic. First, syntactic alterations can lead
to complex regexes, which can harm readability and
reusability~\citep{DavisMCSL19}. On the other hand, semantic changes can defeat
the original purpose of the regex. Therefore, an ideal repair must not only
remove the vulnerability but also maintain syntactic and semantic similarity to
the original regex.

\subsubsection{Repair rate}

We define the \emph{repair rate} as the proportion of vulnerable regexes that
are successfully repaired. We treat ill-formed resulting regexes or zero Jaccard
similarity with the original regexes as failure cases; see
\Cref{sec:semantic-similarity} for details on Jaccard similarity.

We use the ReDoSHunter~\citep{LiCCXPCCC21}, which we chose for its efficiency
and effectiveness. It handles practical regex features such as look-arounds,
back-references, and anchors more reliably than many static analysis tools.
Unlike some other dynamic methods, it consistently produces a result within a
reasonable time. For our evaluation, we classify any regex as invulnerable if
the tool cannot identify a vulnerability within our one-minute timeout
threshold.

\subsubsection{Syntactic similarity}
\label{sec:syntactic-similarity}

We quantify syntactic similarity between an original regex~$s$ and a repaired
regex~$t$ using two metrics. First, the relative length increase~(RLI) measures
the preservation of the regex's size:
\begin{align*}
    \text{RLI}(s, t) = \frac{|t|-|s|}{|s|},
\end{align*}
An RLI value close to zero indicates that the regex's length has been
maintained. We do not define the RLI for ill-formed regexes, as there is no
meaningful reference point~(worst-case value).

The normalized Levenshtein similarity~(NLS) measures the number of the
character-level structural changes:
\begin{align*}
    \text{NLS}(s, t) = 1 - \frac{\text{LD}(s, t)}{\max(|s|, |t|)},
\end{align*}
where $\text{LD}(s, t)$ is the Levenshtein distance, representing the minimum
number of single-character insertions, deletions or substitutions required to
change $s$ into $t$. We give zero NLS scores for ill-formed results.

\subsubsection{Sematic similarity}
\label{sec:semantic-similarity}

For each regex~$s$, we first generate a sample language~$\hat L(s)$ to measure
the semantic similarity between regexes. For each regex, we create their sample
language of size 100 using \texttt{xeger}
package,\footnote{\url{https://pypi.org/project/xeger/}} which perform a random
walk on the regex and generate a sample string matched by the regex. We allow
duplicates to ensure that regexes with sparse languages~(matching few strings)
and those with dense languages~(matching many) are weighted equally in our
analysis.

Given an original regex~$s$ and a repaired regex~$t$, we evaluate their semantic
similarity by counting true positives~(TP), false positives~(FP), true
negatives~(TN), and false negatives~(FN). These metrics are calculated over the
union~$\hat L(s) \cup \hat L(t)$ of their samples. A string is considered a
positive instance if the original regex~$s$ matches it.

Using these counts, we can define standard evaluation metrics such as precision
and recall:
\begin{align*}
    \text{Precision} = \frac{\text{TP}}{\text{TP} + \text{FP}},
    \quad
    \text{Recall} = \frac{\text{TP}}{\text{TP} + \text{FN}}.
\end{align*}
High precision corresponds to soundness; it indicates that strings matched by
the repaired regex~$t$ are also likely to matched by the original regex~$s$.
Conversely, high recall corresponds to completeness; it suggests that strings
matched by the original regex~$s$ are also likely to be captured by the repaired
regex~$t$.

\begin{table*}[htb]
\centering
\begin{tabularx}{\textwidth}{
lX cc cc ccc
}
\toprule
\multirow{2}{*}[-2pt]{Method} &
\multirow{2}{*}[-2pt]{Model} &
\multirow{2}{*}[-2pt]{W.F.} &
\multirow{2}{*}[-2pt]{Repair} &
\multicolumn{2}{c}{Syntactic sim.} &
\multicolumn{3}{c}{Semantic similarity} \\
\cmidrule(r){5-6}
\cmidrule{7-9}
& & & &
RLI~($\downarrow$) & NLS & Jaccard & Prec. & Recall \\
\midrule
\multirow{2}{*}{Baseline}
& RegexScalpel &
\bf 100.00 & 43.60 & 100.53 & 69.60 & 78.71 & \bf 99.91 & 78.77 \\
& Remedy &
46.10 & 26.00 & 102.94 & 21.23 & 12.66 & 37.30 & 13.64 \\
\cmidrule{1-9}
\multirow{9}{*}[-2pt]{LRR~(ours)}
& Qwen3-8B~(NR)&
98.60 & 51.60 & 79.75 & 69.93 & 79.19 & 92.53 & 80.83 \\
& Llama3.1-8B &
96.80 & 50.00 & 119.97 & 58.52 & 73.06 & 85.08 & 74.41 \\
& Gemma3-12B &
99.40 & 47.80 & 63.27 & \bf 76.68 & 83.10 & 93.27 & 84.87 \\
& Phi4-14B &
99.30 & \bf 59.00 & 100.43 & 63.37 & 84.22 & 95.78 & 85.93 \\
\cmidrule{2-9}
& Qwen3-8B &
97.90 & 51.00 & 68.79 & 73.33 & 83.03 & 94.47 & 84.23 \\
& DS-R1-Qwen-7B &
80.70 & 45.60 & 72.44 & 59.34 & 63.86 & 75.03 & 64.86 \\
& GPT-oss-20B~(low) &
91.00 & 47.50 & 63.02 & 69.41 & 78.89 & 88.60 & 79.64 \\
& GPT-oss-20B~(med.) &
96.80 & 47.40 & \bf 61.11 & 74.45 & 84.18 & 94.68 & 84.82 \\
& GPT5 &
99.70 & 49.20 & 62.91 & 75.57 & \bf 89.77 & 98.96 & \bf 90.24 \\

\bottomrule
\end{tabularx}
\caption{\label{tab:overall-experimental-results}
    A performance evaluation of two baseline methods~(RegexScalpel and Remedy)
    against our LRR framework using various LLMs. Qwen3-8B~(NR) represents the
    model operating in the non-reasoning mode. The W.F. column denotes the ratio
    of well-formed output regexes that are compilable. All scores are denoted in
    percent~(\%) and best scores are marked in bold for each metric.
}
\end{table*}

The Jaccard similarity provides a unified view of these metrics, measuring the
overlap between the two regexes:
\begin{align*}
    \text{Jaccard similarity} = \frac{\text{TP}}{\text{TP}+\text{FP}+\text{FN}}.
\end{align*}
A high Jaccard similarity indicates a strong overall agreement between the
original and repaired regexes, where both high precision and high recall are
required. Also, we give zero precision, recall and Jaccard similarity scores for
ill-formed results as well as NLS.

We opt for Jaccard similarity over the F1 score because it better reflects the
set-based nature of languages, directly measuring the overlap between two sets
of strings. Since Jaccard similarity and the F1 score are monotonically related,
they preserve the same ranking, ensuring our choice does not alter the relative
performance assessment.

This entire semantic analysis is conducted using Python's \texttt{Lib/re}
module. While some prior work, like RegexScalpel, was evaluated against a Java
matcher, which only has minor difference between them~\citep{DavisMCSL19}.

\subsection{Dataset}
\label{sec:dataset}

Our evaluation dataset comes from the Polyglot corpus~\citep{DavisMCSL19}, whose
divergency ensures our findings are generalizable and not biased toward a
specific language field. We first identify regexes vulnerable to ReDoS attacks
using the ReDoSHunter~\citep{LiCCXPCCC21} from the Polyglot corpus. We then
filter these results to include only regexes whose length is less than 1,024 to
meet our context size requirements. From this filtered set, we constructed our
dataset of 1,000 regexes to analyze the performance regex repair tools.

We select a balanced sample consisting of two distinct groups: 500~regexes that
result in invulnerable patterns and 500~regexes where the tool fails to complete
within one minute or where the resulting regexes remain vulnerable. We note that
the first group, despite yielding invulnerable results, does not guarantee a
successful repair since it includes cases with zero Jaccard similarity. We
provide detailed in Appendix~\ref{app:failure-regex-scalpel}.

This balanced dataset allows us to not only measure our method's ability to fix
difficult, unresolved regexes but also to understand the structural and semantic
changes that occur during the repair process. By analyzing both successful and
failed repairs, we evaluate LRR's performance and demonstrate its capability to
overcome the limitations of existing rule-based approaches.

\section{Experimental results and analysis}
\label{sec:result-analysis}

The overall results of our performance evaluation are presented in
\Cref{tab:overall-experimental-results}; the full version is provided in
Appendix~\ref{app:full-result}. The table shows the performance increase
obtained by LRR, which incorporates both RegexScalpel and LLM variants. Among
the LRR variants, the top four rows represent results of non-reasoning models,
while the bottom five rows represent reasoning models.

\subsection{Baseline comparisons}
\label{ssec:baseline-comparison}

We first examine the primary baseline, RegexScalpel. The method shows a perfect
score for well-formed ratio but is limited by a repair rate 43.60\%, failing to
resolve a majority of the vulnerabilities. We present such examples that are
then resolved by LRR in \Cref{ssec:case-study}. RegexScalpel often results in
longer regexes, increasing the regex length by approximately a factor of two. We
include analysis on the secondary baseline, Remedy, due to its low well-formed
ratio, which make unsuitable for real-world usage; we discuss this
impracticality in Appendix~\ref{app:failure-remedy}.

In contrast, LRR surpasses the repair rates by 6.30\%p on average, increasing
the rate from 43.60\% to 49.90\%. Most notably, LRR with Phi4 demonstrates the
best repair rate~59.00\%, achieving a 15.40\%p increase over RegexScalpel. This
highlights the effectiveness of combining symbolic analysis with
generalizability of LLMs.

We observe a clear trade-off when applying reasoning: the non-reasoning LLM,
Phi4, improves the repair rate by 8\%p over the reasoning counterpart, Qwen3,
from 51.00\% to 59.00\%. However, this gain is accompanied by a 31.64\%p
increase in RLI from 68.79\% to 100.43\%. This empirically proves that
vulnerability repair often requires an input regex to be much longer. We detail
this analysis in \Cref{ssec:syntactic-similarity}.

\subsection{Ablation studies}
\label{ssec:ablation}

We conduct an ablation study of LRR to evaluate the individual contributions of
our prompt components: vulnerability localization and 5-shot CoT prompting. The
analysis in \Cref{tab:ablation} uses Phi4 and Qwen3 as representatives of the
best-performing non-reasoning and reasoning LLMs, respectively.

\begin{table}[hbt]
\centering
\begin{tabularx}{\linewidth}{Xccc}
\toprule
Model &
Repair &
NLS &
Jaccard \\
\midrule
RegexScalpel &
43.60 & 30.40 & 78.71 \\
\cmidrule{1-4}
Phi4-14B &
59.00 & 63.37 & 84.22 \\
\cdashline{1-4} \noalign{\vskip 3pt}
(-) localization &
57.80 & 65.72 & 83.61 \\
(-) 5-shot &
49.70 & 72.40 & 76.91 \\
(-) local. \& 5-shot &
49.00 & 26.95 & 78.88 \\
\cmidrule{1-4}
Qwen3-8B &
51.00 & 73.33 & 83.03 \\
\cdashline{1-4} \noalign{\vskip 3pt}
(-) localization &
46.60 & 77.10 & 83.61 \\
(-) 5-shot &
47.60 & 75.39 & 79.11 \\
(-) local. \& 5-shot &
47.30 & 75.74 & 79.58 \\
\bottomrule
\end{tabularx}
\caption{\label{tab:ablation}
    Ablation study on our prompt components: vulnerability localization and
    5-shot CoT prompting.
}
\end{table}

The results reveal a distinct dependency for model types. For the non-reasoning
Phi4 model, the 5-shot CoT prompting is the most critical component. This
removal causes a significant drop in the repair rate, suggesting that the models
rely on the examples to learn the procedural knowledge of the repair task. In
contrast, the reasoning-based Qwen3 model is most impacted by removing
localization. This indicates that it requires precise guidance on where to apply
its reasoning.

Consequently, the study confirms that the two prompt components are synergistic
and complementary. Localization tells the model what to fix, which is especially
crucial for reasoning models, while the 5-shot CoT prompting teaches the models
how to fix, specifically effective for guiding non-reasoning models. By
providing both a focused problem scope and a clear solution procedure, LRR is
robustly effective across different LLM.

\subsection{Analysis on syntactic similarity}
\label{ssec:syntactic-similarity}

\begin{figure}[htb]
\centering
\begin{subfloat}[Relative length increase]{\label{fig:relative-legnth-increase}
    \includegraphics[width=0.95\linewidth]{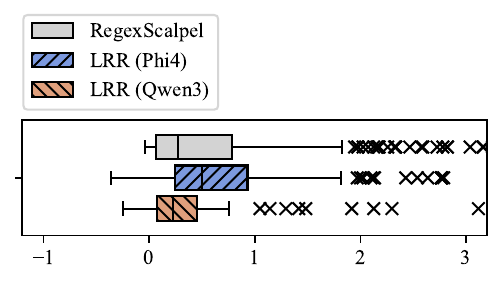}
}
\end{subfloat}
\begin{subfloat}[Normalized Levenshtein similarity]{
\label{fig:normalized-levenshtein-similarity}
    \includegraphics[width=0.95\linewidth]{
        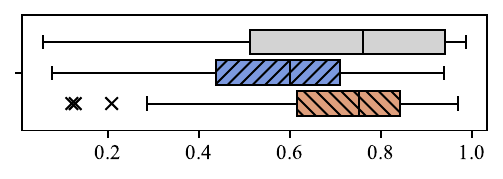
    }
}
\end{subfloat}
\caption{\label{fig:syntactic-similarity}
    Box-and-whisker plots of the RLI~(top) and NLS~(bottom) of successfully
    repaired cases for each method. For the LRR method, we exclude all cases
    successfully repaired by the baseline RegexScalpel to isolate and highlight
    the unique contribution of LLMs.
}
\end{figure}

\Cref{fig:syntactic-similarity} illustrates the distribution of the two
syntactic similarity scores; each box represents the inter-quartile range~(IQR),
with the whiskers extending up to l.5 times the IQR. Both RegexScalpel and Phi4
exhibit a higher RLI compared to the Qwen3 variant. This disparity suggests that
RegexScalpel and Phi4 have a higher tendency to add constructs like look-aheads
during the repair process, whereas the Qwen3 model shows a relatively lower
propensity for such additions. This lower propensity implies that the Qwen3
model performs repair in a relatively more stable manner.

This stability of the reasoning model stems from a tendency to apply the
simplest strategy with the minimal modifications first, and then it stops when
the vulnerability is resolved. On the other hand, the non-reasoning models often
apply all applicable rules at once. Considering that most ReDoS vulnerability
resolution strategies involve increasing the regex length, the unnecessary
application of additional strategies contributed to the reduced syntactic
similarity of the non-reasoning model. We provide detailed examples illustrating
these behavioral differences between the reasoning and non-reasoning models in
Appendix~\ref{app:syntactic-similarity}.

The distribution of NLS shows an inverse trend compared to the RLI scores. Given
that an RLI close to zero indicates the repaired regex maintains its original
length, this inverse relationship suggests that repair is primarily achieved by
adding characters to the regex.

\subsection{Analysis on semantic similarity}
\label{ssec:semantic-similarity}

\begin{figure}[htb]
\centering
\begin{subfloat}[LRR with non-reasoning LLM; Phi4-14B]{\label{fig:case-study-0}
    \includegraphics[width=0.95\linewidth]{
        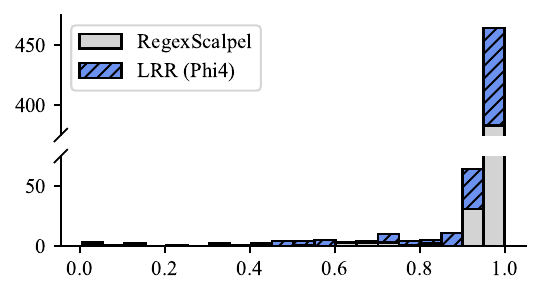
    }
}
\end{subfloat}
\begin{subfloat}[LRR with reasoning LLM; Qwen3-8B]{\label{fig:case-study-1}
    \includegraphics[width=0.95\linewidth]{
    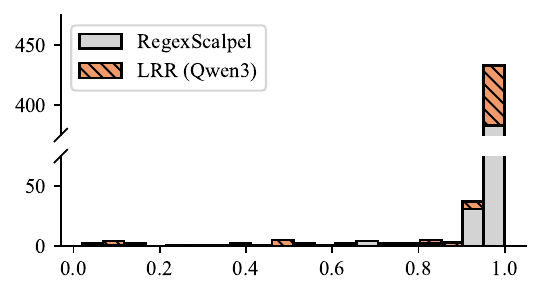
    }
}
\end{subfloat}
\caption{\label{fig:jaccard}
    Histograms illustrating the distribution of Jaccard Similarity scores for
    successfully repaired regexes.
}
\end{figure}

The primary goal of LRR is to produce a robust and semantically equivalent
repair. \Cref{fig:jaccard} illustrates the distribution of successful repair
scores and Jaccard similarity for LRR with Phi4 and Qwen3 as strong
non-reasoning and reasoning LLMs, respectively. The analysis reveals a trade-off
between the quantity of repairs and their semantic correctness.

LRR with Phi4 successfully captures and repairs vulnerable regexes that
RegexScalpel fails, shown in \Cref{fig:case-study-0}. However, the distribution
of these similarity scores, while strong, is less concentrated at the
full-score~(1.0) mark than its reasoning-based counterpart.
\Cref{fig:case-study-1} illustrates that Qwen3 produces repairs of more reliable
semantic quality. Although Phi4 shows the most effective repair rate, a
significantly large proportion of successful repairs by Qwen3 achieve
near-perfect Jaccard similarity. This suggests that the reasoning model is more
capable of maintaining semantic correctness while repairing vulnerabilities.

Perhaps, its high precision~(99.91\%) is consistent with the claim by
\citet{LiSXCLLCCLX22} that their heuristic is designed to never expand the
pattern's languages, thereby preventing false positives of the repaired pattern.
In contrast, the LRR methodology exhibits a slightly reduced precision, by
aiming to eliminate a greater number of vulnerabilities.

\subsection{Case-study}
\label{ssec:case-study}

We analyze cases where LRR succeeds while symbolic baselines such as
RegexScalpel and Remedy fails to reveal why our approach improves the overall
repair rate. The following examples highlight the key advantages of LLMs,
specifically their ability to perform contextual reasoning. This allows the LLMs
to infer the semantic intent of given regexes---what the expression is meant to
match---rather than restricting the analysis solely on the syntactic structure.
This context-aware understanding enables LRR to recognize and repair the regexes
that pure rule-based approaches cannot address.

\begin{examplebox}[title={Examples: Contextual Repairs}]
\small
\hphantom{Repaired:}\llap{Original:}\\
\hspace*{12pt}\verb`<\?(=|php)(.+?)\?>`\\
Repaired:\\
\hspace*{12pt}\verb`<\?(=|php)([^\?>]+?)\?>`

\vspace{-3pt}\hrulefill\vspace{2pt}

\hphantom{Repaired:}\llap{Original:}\\
\hspace*{12pt}\verb`<a href="([^"]+)">(.+?)</a>`\\
Repaired:\\
\hspace*{12pt}\verb`<a href="([^"]+)">([^<]+)</a>`

\vspace{-3pt}\hrulefill\vspace{2pt}

\hphantom{Repaired:}\llap{Original:}\\
\hspace*{12pt}\verb`@@ \-(\d+),?(\d+)? \+(\d+),?(\d+)? @@`\\
Repaired:\\
\hspace*{12pt}\verb`@@ \-(\d{1,3}(?:,\d{2})?) \+(\d{1,3}... @@`
\end{examplebox}

This capability to understand the context of regexes is especially evident in
domain-specific formats. For instance from the above example, the LLM recognizes
the \texttt{@@}\ldots\texttt{@@} signature as a git diff header and infers that
\verb`\d+` represents line counts. Consequently, the LLMs constrains the pattern
to a more plausible range. This demonstrates the effectiveness of contextual
understanding of LLMs in our LRR framework that moves beyond a rule-based
approach.

\section{Conclusion}
\label{sec:conclusion}

We propose and validate a hybrid framework that addresses the core challenge of
automated ReDoS repair. Traditional methods are fundamentally limited:
synthesis-based approaches require large data dependency, while rule-based
systems lack the generalizability.

Our framework overcomes these limitations by integrating the localize-and-fix
strategy with the advanced generative capabilities of LLMs. The symbolic module
performs precise vulnerability localization, identifying the vulnerable
sub-pattern. The LLM leverages this focused, actionable context to perform the
repair, effectively transforming the broad, complex task into a manageable one.

Experiments confirm this approach successfully repairs complex patterns that
traditional rule-based systems fail to address. The resulting regexes not only
eliminate the vulnerability but also show high syntactic and semantic
similarity, ensuring the patterns remain both safe and practical.

In conclusion, our work demonstrates a robust and effective method for automated
ReDoS vulnerability repair, offering a promising path for developing more
adaptable and scalable solutions by combining precise analytical tools with
powerful generative models.

\section*{Limitations}

The LRR framework, while demonstrating strong practical efficacy in automated
ReDoS repair, operates under two primary, empirical constraints.

First, the symbolic localization module, which effectively scopes the repair
task, relies on established heuristics derived from existing rule-based repair
methods. This dependence means LRR's effectiveness is constrained by the
knowledge of the underlying analyzer, potentially limiting its ability to
precisely localize vulnerability patterns that fall outside of these known
anti-patterns. If the localization module fails to pinpoint the vulnerable
sub-segment, the LLM is forced to analyze the entire, unconstrained regex, which
negates the core efficiency benefit of the localize-and-fix strategy.

Second, the validation of our repair success is empirical, not a formal,
provable guarantee of safety, as it relies on the dynamic detection tool
ReDoSHunter. The final invulnerability status is defined by the tool's failure
to find an attack string within a one-minute timeout threshold. Moreover, given
that the definition of a ReDoS vulnerability is often context-dependent, varying
across different execution environments and regex engines, there is no single,
universally consistent standard for definitive
invulnerability~\cite{HassanALDS23}. Thus, the reported repair rate is best
viewed as robust evidence of practical safety relative to the fidelity of the
specific detection mechanism employed.


\bibliography{references}

\newpage

\appendix

\section{Experimental details}
\label{app:experimental-details}

During our experiments, LLM inferences are executed on systems running the Rocky
Linux 9.4 operating system, leveraging an NVIDIA RTX A6000 graphic processing
unit. RegexScalpel and ReDoS{-}Hunter are allocated to a Rocky Linux 8.10
system, which is powered by an AMD Ryzen Threadripper 3960X 24-core
processor~(48 threads). Regarding the software stack, both RegexScalpel and
ReDoSHunter are built using Java openjdk 25, while the core implementation of
the LRR framework itself was written in Python 3.12.

\section{Prompts}
\label{app:prompts}

The following are prompts employed to configure the LLMs within the LRR
framework.

\begin{promptbox}[title=Basic prompt]\small
You are an expert Application Security Analyst specializing in regular
expression security.

$\cdots$

Your entire output should be only the final, corrected regular expression,
enclosed in a code block.

For example, if the input is: \verb`(?:a+)+`

your output must be: \verb`a+`

Analyze the following pattern: \verb`{pattern}`

\end{promptbox}

\begin{promptbox}[title=5-shot prompt]\small
\verb`{Basic prompt}`

\vspace{5pt}

Here is an example of the repairing process.

\verb`{Five examples of reparing process}`

\vspace{5pt}

Analyze the following pattern: \verb`{pattern}`

\end{promptbox}

\begin{promptbox}[title=Localization prompt]\small
\verb`{Basic prompt}`

\vspace{5pt}

You must focus on the following subpatterns: \verb`{subpatterns}`. At least one
of these subpattern causes catastrophic backtracking.

\vspace{5pt}

Analyze the following pattern; you have to answer the repaired version of the
entire pattern, not only a repaired subpattern: \verb`{pattern}`

\end{promptbox}

\begin{promptbox}[title=LRR prompt]\small
\verb`{Basic prompt}`

\vspace{5pt}

Here is an example of the repairing process.

\verb`{Five examples of repairing process}`

\vspace{5pt}

You must focus on the following subpatterns: \verb`{subpatterns}`. At least one
of these subpattern causes catastrophic backtracking.

\vspace{5pt}

Analyze the following pattern; you have to answer the repaired version of the
entire pattern, not only a repaired subpattern: \verb`{pattern}`

\end{promptbox}

\section{Failure of baseline methods}
\label{app:failure-baselines}

In this section, we analysis failure cases of rule-based baseline methods.

\subsection{RegexScalpel}
\label{app:failure-regex-scalpel}

RegexScalpel~\citep{LiSXCLLCCLX22} uses negative look-aheads~\verb`(?!...)`,
which assert that the current position in the string is not followed by a
substring that matches the pattern~\verb`r`, to remove overlapping patterns.
This effectively removes the ReDoS vulnerability by eliminating unambiguous
choices during matching. However, if the pattern~\verb`r` is chosen with too
broad, the resulting overall pattern becomes too narrow. As a result, of the 500
ReDoS-invulnerable regexes generated by RegexScalpel, 39 exhibited a Jaccard
similarity of zero when compared to their original patterns.

The following repair example results in an empty language pattern, which cannot
match any strings; refer the rule~$\tau_{33}$ from Table~8 in
\citet{LiSXCLLCCLX22}.

\begin{examplebox}[title={Failure example of RegexScalpel}]\small
\hphantom{Repaired:}\llap{Original:}\\
\hspace*{12pt}\verb`<(named-content.*?)>`\\
Repaired:\\
\hspace*{12pt}\verb`(?!.*?)<(named-content.*?)>`
\end{examplebox}

\subsection{Remedy}
\label{app:failure-remedy}

As shown in \Cref{tab:overall-experimental-results}, Remedy~\citep{ChidaT22}
exhibits a low ratio of well-formed repaired regexes. This occurs because Remedy
does not fully support common special symbols frequently found in real-world
patterns, such as character classes like \verb`\d`~(digits) and \verb`\w`~(word
characters).

The following failure example illustrates this limitation: the presence of the
\verb`\d` symbol, which represents digits, causes Remedy to fail compilation or
raise an internal error, resulting in an ill-formed output and contributing to
its low overall performance score.

\begin{examplebox}[title={Failure example of Remedy}]\small
Original:\\
\verb`^POINT\\((-?\\d+\\.?\\d*) (-?\\d+\\.?\\d*)\\$`

$\hookrightarrow$ \verb`Remedy raises an error`
\end{examplebox}

\section{Full experimental results}
\label{app:full-result}

\Cref{tab:overall-experimental-results} from Section~\ref{sec:result-analysis}
demonstrates our main results. This section provides the full results. of LRR
across all our base LLMs.
Tables~\ref{tab:full-nonreasoning-experimental-result}
and~\ref{tab:full-reasoning-experimental-result} present the performance of LRR
with non-reasoning and reasoning LLMs, respectively. For each model, we report
the performance of our full framework, which combines vulnerability localization
with 5-shot CoT prompting. We also include results from three variants:
\mbox{(-)}~localization, \mbox{(-)}~5-shot, and \mbox{(-)}~localization \&
5-shot. to demonstrate the individual contributions of each component. Likewise
to Section~\ref{sec:result-analysis}, the evaluation metrics cover the rate of
well-formed outputs, repair success rate, syntactic similarity~(RLI, NLS), and
semantic similarity~(Jaccard, precision, recall). The results confirm that our
observation in \Cref{ssec:ablation} consists over LLMs.

\begin{table*}
\centering
\begin{tabularx}{\textwidth}{Xccccccc}
\toprule
\multirow{2}{*}[-2pt]{Model} &
\multirow{2}{*}[-2pt]{Well-formed} &
\multirow{2}{*}[-2pt]{Repair} &
\multicolumn{2}{c}{Syntactic similarity} &
\multicolumn{3}{c}{Semantic similarity} \\
\cmidrule(r){4-5}
\cmidrule{6-8}
& & &
RLI & NLS & Jaccard & Precision & Recall \\
\midrule
Qwen3-8B~(NR) &
98.60 & 51.60 & 79.75 & 69.93 & 79.19 & 92.53 & 80.83 \\
\cdashline{1-8} \noalign{\vskip 3pt}
(-) localization &
98.70 & 50.70 & 66.29 & 73.88 & 80.00 & 93.16 & 81.14 \\
(-) 5-shot &
99.40 & 50.40 & 51.98 & 75.20 & 77.08 & 91.14 & 79.03 \\
(-) local. \& 5-shot &
99.20 & 49.30 & 52.61 & 75.16 & 77.36 & 91.11 & 79.32 \\

\cmidrule{1-8}

Llama3.1-8B &
96.80 & 50.00 & 119.97 & 58.52 & 73.06 & 85.08 & 74.41 \\
\cdashline{1-8} \noalign{\vskip 3pt}
(-) localization &
98.00 & 48.80 & 103.20 & 68.79 & 79.32 & 91.21 & 80.73 \\
(-) 5-shot &
95.60 & 48.50 & 58.26 & 72.25 & 75.92 & 87.93 & 78.06 \\
(-) local. \& 5-shot &
95.80 & 48.10 & 61.61 & 73.22 & 75.93 & 89.06 & 77.85 \\

\cmidrule{1-8}

Gemma3-12B &
99.40 & 47.80 & 63.27 & 76.68 & 83.10 & 93.27 & 84.87 \\
\cdashline{1-8} \noalign{\vskip 3pt}
(-) localization &
99.40 & 46.10 & 54.69 & 80.18 & 84.96 & 94.17 & 86.38 \\
(-) 5-shot &
99.40 & 46.10 & 54.00 & 77.79 & 81.49 & 91.45 & 83.61 \\
(-) local. \& 5-shot &
99.60 & 45.90 & 53.80 & 78.91 & 82.24 & 92.22 & 84.22 \\

\cmidrule{1-8}

Phi4-14B &
99.30 & 59.00 & 100.43 & 63.37 & 84.22 & 95.78 & 85.93 \\
\cdashline{1-8} \noalign{\vskip 3pt}
(-) localization &
99.60 & 57.80 & 93.94 & 65.72 & 83.61 & 95.13 & 85.44 \\
(-) 5-shot &
97.90 & 49.70 & 62.24 & 72.40 & 76.91 & 89.55 & 79.51 \\
(-) local. \& 5-shot &
97.40 & 49.00 & 62.38 & 73.05 & 78.88 & 90.58 & 81.31 \\

\bottomrule
\end{tabularx}
\caption{
\label{tab:full-nonreasoning-experimental-result}
    Full experimental results of LRR on non-reasoning large language models.
}
\end{table*}

\begin{table*}
\centering
\begin{tabularx}{\textwidth}{Xccccccc}
\toprule
\multirow{2}{*}[-2pt]{Model} &
\multirow{2}{*}[-2pt]{Well-formed} &
\multirow{2}{*}[-2pt]{Repair} &
\multicolumn{2}{c}{Syntactic similarity} &
\multicolumn{3}{c}{Semantic similarity} \\
\cmidrule(r){4-5}
\cmidrule{6-8}
& & &
RLI & NLS & Jaccard & Precision & Recall \\
\midrule
Qwen3-8B &
97.90 & 51.00 & 68.79 & 73.33 & 83.03 & 94.47 & 84.23 \\
\cdashline{1-8} \noalign{\vskip 3pt}
(-) localization &
98.20 & 46.60 & 57.24 & 77.10 & 83.61 & 94.62 & 84.36 \\
(-) 5-shot &
97.90 & 47.60 & 53.61 & 75.39 & 79.11 & 91.64 & 80.84 \\
(-) local. \& 5-shot &
97.40 & 47.30 & 54.51 & 75.74 & 79.58 & 91.41 & 80.76 \\

\cmidrule{1-8}

DS-R1-Qwen-7B &
80.70 & 45.60 & 72.44 & 59.34 & 63.86 & 75.03 & 64.86 \\
\cdashline{1-8} \noalign{\vskip 3pt}
(-) localization &
74.20 & 44.80 & 76.62 & 55.55 & 61.92 & 71.20 & 62.55 \\
(-) 5-shot &
75.20 & 45.60 & 72.93 & 55.54 & 60.66 & 70.80 & 61.51 \\
(-) local. \& 5-shot &
73.50 & 45.30 & 74.01 & 54.14 & 60.76 & 69.73 & 61.62 \\

\cmidrule{1-8}

GPT-oss-20B~(low) &
91.00 & 47.50 & 63.02 & 69.41 & 78.89 & 88.60 & 79.64 \\
\cdashline{1-8} \noalign{\vskip 3pt}
(-) localization &
89.90 & 46.40 & 62.79 & 70.27 & 79.55 & 88.22 & 80.04 \\
(-) 5-shot &
98.30 & 46.70 & 57.61 & 76.41 & 85.21 & 95.14 & 86.28 \\
(-) local. \& 5-shot &
97.70 & 45.80 & 56.49 & 78.71 & 86.63 & 95.49 & 87.14 \\

\cmidrule{1-8}

GPT-oss-20B~(med.) &
96.80 & 47.40 & 61.11 & 74.45 & 84.18 & 94.68 & 84.82 \\
\cdashline{1-8} \noalign{\vskip 3pt}
(-) localization &
85.90 & 44.90 & 63.64 & 67.87 & 76.62 & 84.64 & 76.94 \\
(-) 5-shot &
97.10 & 46.60 & 59.17 & 75.63 & 83.18 & 94.62 & 83.71 \\
(-) local. \& 5-shot &
98.10 & 45.60 & 55.75 & 79.39 & 87.65 & 96.79 & 88.08 \\

\cmidrule{1-8}

GPT-5 &
99.70 & 49.20 & 62.91 & 75.57 & 89.77 & 98.96 & 90.24 \\
\cdashline{1-8} \noalign{\vskip 3pt}
(-) localization &
99.90 & 48.30 & 57.77 & 79.33 & 90.54 & 99.56 & 90.73 \\
(-) 5-shot &
99.50 & 47.30 & 60.30 & 77.30 & 89.54 & 98.98 & 89.80 \\
(-) local. \& 5-shot &
100.00 & 46.20 & 56.17 & 81.04 & 91.33 & 99.48 & 91.53 \\

\bottomrule
\end{tabularx}
\caption{
\label{tab:full-reasoning-experimental-result}
    Full experimental results of LRR on reasoning large language models.
}
\end{table*}

\section{Behavioral differences between models}
\label{app:syntactic-similarity}

The following is selected examples which highlight the difference between
non-reasoning (Phi4) and reasoning~(Qwen3) models. They utilize two strategies:
\begin{enumerate}[noitemsep]
    \item[(a)] match-all symbol~`\texttt{.}' \\
    $\to$ character class~`\verb`[...]`'
    \item[(b)] unbounded repetition~`\texttt{*}' and `\texttt{+}' \\
    $\to$ bounded repetition~`\verb`{n,m}`'
\end{enumerate}
This distinct strategic behavior results in the non-reasoning model applying
(a), (a, b), and (a, b) sequences, while the reasoning model uses (a), (b), and
(b), yielding a more syntactically similar repair.

\begin{examplebox}[title={Repair with different syntactics}]\small
Original:\\
\hspace*{12pt}\verb`^\n*(.*?)\s*$`\\
Repaired by Phi4:\\
\hspace*{12pt}\verb`^\n{0,1000}(.{0,1000})\s{0,1000}$`\\
Repaired by Qwen3:\\
\hspace*{12pt}\verb`^\n*[\s\S]*\s*$`

\vspace{-3pt}\hrulefill\vspace{2pt}

Original:\\
\hspace*{12pt}\verb`image:: (.*\.png)`\\
Repaired by Phi4:\\
\hspace*{12pt}
\verb`image:: ([\u0000-...-\u007f]{0,1000}\.png)`\\
Repaired by Qwen3:\\
\hspace*{12pt}\verb`image:: (.{0,500}\.png)`

\vspace{-3pt}\hrulefill\vspace{2pt}

Original:\\
\hspace*{12pt}\verb`/i18n/(.+)\.json$`\\
Repaired by Phi4:\\
\hspace*{12pt}\verb`/i18n/([\u0000-\uFFFF]{1,100})\.json`\\
Repaired by Qwen3:\\
\hspace*{12pt}\verb`/i18n/(.{0,500})\.json$`

\end{examplebox}

\end{document}